\documentclass[letterpaper]{article}

\PassOptionsToPackage{table}{xcolor}
\usepackage[preprint]{aaai2027}
\usepackage[hyphens]{url}
\usepackage{graphicx}
\usepackage{natbib}
\usepackage{caption}
\usepackage{amsmath}
\usepackage{amsfonts}
\usepackage{booktabs}
\usepackage{algorithm}
\usepackage{algpseudocode}
\usepackage{hyperref}

\newcommand{\method}{\textsc{GramLoop}}
\newcommand{\up}{\ensuremath{\uparrow}}
\newcommand{\down}{\ensuremath{\downarrow}}
\newcommand{\best}[1]{\textbf{#1}}
\newcommand{\deltaPosColor}{green!45!black}
\newcommand{\deltaNegColor}{red!55!black}

\title{
GramLoop: Training-Free Gram-Gated Replay for Robust Dense Prediction}
\author{
Yang Chen\textsuperscript{\rm 1},
Canyu Shen\textsuperscript{\rm 2},
Xinzhe Rao\textsuperscript{\rm 1},
Yuanyi Yan\textsuperscript{\rm 1}\\
Yunlu Chen\textsuperscript{\rm 3},
Meng Tang\textsuperscript{\rm 4},
Teng Long\textsuperscript{\rm 5},
Vincent Tao Hu\textsuperscript{\rm 1}
}
\affiliations{
\textsuperscript{\rm 1}Huazhong University of Science and Technology\\
\textsuperscript{\rm 2}Tongji University\\
\textsuperscript{\rm 3}King Abdullah University of Science and Technology\\
\textsuperscript{\rm 4}University of California, Merced\\
\textsuperscript{\rm 5}University of Amsterdam
}

\begin{document}

\maketitle

\begin{abstract}
  We aim to improve frozen DINOv3 dense-prediction models under distribution
shift by adding inference computation inside the visual backbone, without
changing model weights, task adapters, or prediction heads. The challenge is
that repeated transformer-block computation must refine dense features without
disrupting the pairwise patch relations that DINOv3 uses to preserve spatial
structure. We introduce \method{}, a training-free framework that replays a
short transformer window and controls each replay through final-layer
cosine-Gram consistency. Each proposal is propagated through the frozen suffix,
measured against the standard DINOv3 trajectory, and accepted through a
patchwise gate at the replay-window endpoint. Across object detection and
semantic segmentation under corruptions, perturbations, and natural shifts,
\method{} improves all five shifted benchmarks over the paired DINOv3 baseline.
On COCO-O, it improves mAP by \(+0.252\) and Effective Robustness by \(+0.250\),
while preserving clean ADE20K performance. Code will be released at
\href{https://github.com/cheyan9/GramLoop}{\texttt{this https URL}}. \end{abstract}

\section{Introduction}

Built on the patch-token architecture introduced by
ViT~\citep{dosovitskiy2021vit}, self-supervised vision transformers have become
general-purpose feature extractors for both image-level recognition and dense
prediction.
DINO~\citep{caron2021dino} showed that self-distillation can produce
semantically organized patch representations without manual labels, while
DINOv2~\citep{oquab2024dinov2} and DINOv3~\citep{simeoni2026dinov3}
substantially scaled this paradigm in model capacity and pretraining data. In
particular, DINOv3 scales self-supervised pretraining to models with up to 7B
parameters and 1.7B unlabeled images, producing frozen features that support
segmentation, object detection, and depth estimation
\citep{simeoni2026dinov3}. This combination of scalability and dense spatial
structure makes DINOv3 an attractive backbone for downstream visual prediction.

Nevertheless, strong performance on clean benchmarks does not guarantee
robustness under challenging acquisition conditions. Common corruptions degrade
semantic-segmentation systems \citep{kamann2021segmentationcorruptions} and
object detectors \citep{michaelis2019robustness}. Real-world-inspired
perturbations expose similar weaknesses in foundation-model segmentation
\citep{schiappa2024robustness}, while natural distribution shifts reduce
detector performance \citep{mao2023cocoo}.
This limitation motivates a complementary question: can the representation
trajectory of a frozen visual backbone be refined at inference time, without
updating either the backbone or its task-specific head?

Looped and recurrent-depth transformers provide a natural mechanism for
allocating additional computation by repeatedly applying shared transformer
blocks. Universal Transformers introduced recurrent self-attention with
adaptive halting \citep{dehghani2019ut}. Subsequent work studied looped
transformers as programmable computers \citep{giannou2023pc}, iterative
in-context learners \citep{yang2024lt}, and latent-reasoning models
\citep{saunshi2025lt}. More recently, training-free looped transformers
retrofit damped block replay onto frozen language models and show that naive
block reapplication can be harmful \citep{chen2026looped}. However, it remains
unclear how such replay should be controlled spatially when applied to frozen
visual features for dense prediction under distribution shifts: replay proposes
changes to every token, although only some may preserve the spatial structure
of the frozen backbone.

DINOv3 provides a useful structural signal through its Gram-anchoring objective,
which preserves pairwise relations among patch features during large-scale
pretraining \citep{simeoni2026dinov3}. We hypothesize that replay proposals
causing less downstream Gram drift are more reliable. Based on this hypothesis,
we introduce \method, which propagates each proposal through the frozen suffix,
measures its patchwise final-layer cosine-Gram drift from the standard DINOv3
trajectory, and uses this drift to scale the update at the replay-window
endpoint. High-drift updates are attenuated, while the task adapter and
prediction head remain unchanged. A matched-strength diagnostic confirms lower
replay-induced Gram discrepancy than uniform replay across all 17 COCO-P
perturbations and all 85 severity-specific conditions.

Our contributions are threefold:
\begin{itemize}
    \item We introduce \method{}, a training-free replay framework for frozen
    DINOv3 dense-prediction models. It uses patchwise final-layer
    cosine-Gram drift to selectively control feature updates at an earlier
    replay-window endpoint, without updating the backbone, task adapter, or
    prediction head.

    \item We establish the effectiveness of \method{} across semantic
    segmentation and object detection under controlled corruptions,
    real-world-inspired perturbations, and natural distribution shifts.
    \method{} improves COCO-O, COCO-P, COCO-C, ADE20K-C, and ADE20K-P while
    preserving clean performance.

    \item We provide controlled diagnostics and ablations explaining these
    gains. At matched mean update strength, patchwise Gram gating reduces
    final-layer relation error relative to uniform replay across all 17 COCO-P
    perturbations and 85 severity-specific conditions. A placement ablation
    further identifies final-layer Gram measurement with loop-end updates as
    the strongest evaluated configuration.
\end{itemize}
 \section{Related Work}

\subsection{DINO Features}
DINO learns semantically structured vision-transformer features through
self-distillation without labels \citep{caron2021dino}. DINOv2 scales this
formulation to diverse curated data and billion-parameter models, producing
transferable frozen representations for both image- and pixel-level tasks
\citep{oquab2024dinov2}. DINOv3 further scales pretraining and introduces Gram
anchoring to prevent the degradation of dense patch relations during long
training schedules \citep{simeoni2026dinov3}. Recent work has also used frozen
DINOv3 features with support-derived prototypes and channel-wise Gram
refinement for training-free few-shot segmentation \citep{zakir2026fssdino}. In
contrast, \method{} requires no support set and does not construct class
prototypes: it measures within-image patch-to-patch Gram drift to gate replayed
backbone features before an unchanged dense-prediction head. Despite this broad
transfer, foundation-model segmentation remains sensitive to real-world-inspired
perturbations \citep{schiappa2024robustness}, while object detectors degrade
under natural distribution shifts \citep{mao2023cocoo}. Our work therefore
studies whether DINOv3's own relational structure can control inference-time
feature refinement under these shifts.

\subsection{Test-Time Adaptation}
Test-time adaptation uses unlabeled test inputs to respond to distribution
shift. Test-time training updates model parameters on each test sample through
a self-supervised objective \citep{sun2020ttt}, while Tent minimizes prediction
entropy by updating normalization statistics and affine parameters
\citep{wang2021tent}. CoTTA addresses continual non-stationary test streams
with prediction averaging and stochastic restoration \citep{wang2022cotta}, and
EATA selects reliable samples while regularizing important parameters to reduce
forgetting \citep{niu2022eata}. More recent vision-language methods avoid online
backpropagation: ZERO selects and aggregates confident predictions from
augmented views in a single batched encoder pass \citep{farina2024zero},
whereas SSG reweights visual features and updates a test-time cache using shape
and style guidance \citep{zhou2025ssg}. In contrast, \method{} neither
optimizes parameters nor aggregates output predictions; it refines frozen dense
features through controlled forward replay while leaving the downstream head
unchanged.

\subsection{Looping Transformers}
Recurrence in depth allows a transformer to reuse parameters while increasing
its effective computation. Universal Transformers introduced recurrent
self-attention with adaptive halting \citep{dehghani2019ut}. Deep Equilibrium
Models provide a related weight-tied view by solving for the fixed point of an
effectively infinite-depth transformation \citep{bai2019deq}. Multiscale DEQs
extend implicit recurrence to multiresolution vision features, including
semantic segmentation \citep{bai2020mdeq}. Later looped transformers were
studied as programmable computers \citep{giannou2023pc}, iterative in-context
learners \citep{yang2024lt}, and latent-reasoning models
\citep{saunshi2025lt}. These methods generally train the recurrent structure
end to end. Recent work further learns token-level deciders that selectively
allocate latent iterations in language models \citep{fu2026tah}. In vision,
A-ViT learns token-wise halting depths to allocate input-dependent computation
\citep{yin2022avit}. Training-free looped transformers instead replay a
contiguous block of a frozen language model using damped updates
\citep{chen2026looped}.
\method{} retrofits replay onto a pretrained frozen vision foundation model.
It measures proposal-specific global cosine-Gram drift at the final layer and
uses this relational signal to control each patch-token update, refining the
frozen backbone trajectory without modifying the model parameters or
dense-prediction head.

\subsection{Dense Prediction}
Dense prediction assigns structured outputs to spatial image regions, including
semantic labels for segmentation and categories and boxes for detection. DPT
demonstrated that transformer backbones can support dense depth estimation and
semantic segmentation \citep{ranftl2021dpt}. SETR recast semantic segmentation
as sequence-to-sequence prediction with a pure transformer encoder
\citep{zheng2021setr}, and SegFormer paired a hierarchical transformer encoder
with a lightweight decoder \citep{xie2021segformer}. DETR formulated object
detection as end-to-end set prediction with a transformer encoder-decoder
\citep{carion2020detr}. Modern systems combine general-purpose backbones with
task-specific adapters and decoders, such as ViT-Adapter
\citep{chen2023vitadapter}, Mask2Former \citep{cheng2022mask2former}, and
Plain-DETR \citep{lin2023plaindetr}.
Self-supervised foundation models reduce the label dependence of backbone
pretraining, but their downstream segmentation and detection heads are still
trained with task annotations; the resulting pipeline should therefore not be
described as fully unsupervised. DINOv3 follows this paradigm by keeping the
pretrained backbone frozen while fitting dense-prediction adapters and heads
\citep{simeoni2026dinov3}. Robustness benchmarks further show that
semantic-segmentation systems remain sensitive to common corruptions
\citep{kamann2021segmentationcorruptions} and real-world-inspired perturbations
\citep{schiappa2024robustness}, while object detectors degrade under synthetic
corruptions \citep{michaelis2019robustness} and natural domain shifts
\citep{mao2023cocoo}.
\method{} leaves both the released adapter and task head unchanged and modifies
only the frozen backbone trajectory at inference time, enabling the same
mechanism to support semantic segmentation and object detection.
 \begin{figure*}[t]
  \centering
  \includegraphics[width=0.99\textwidth]{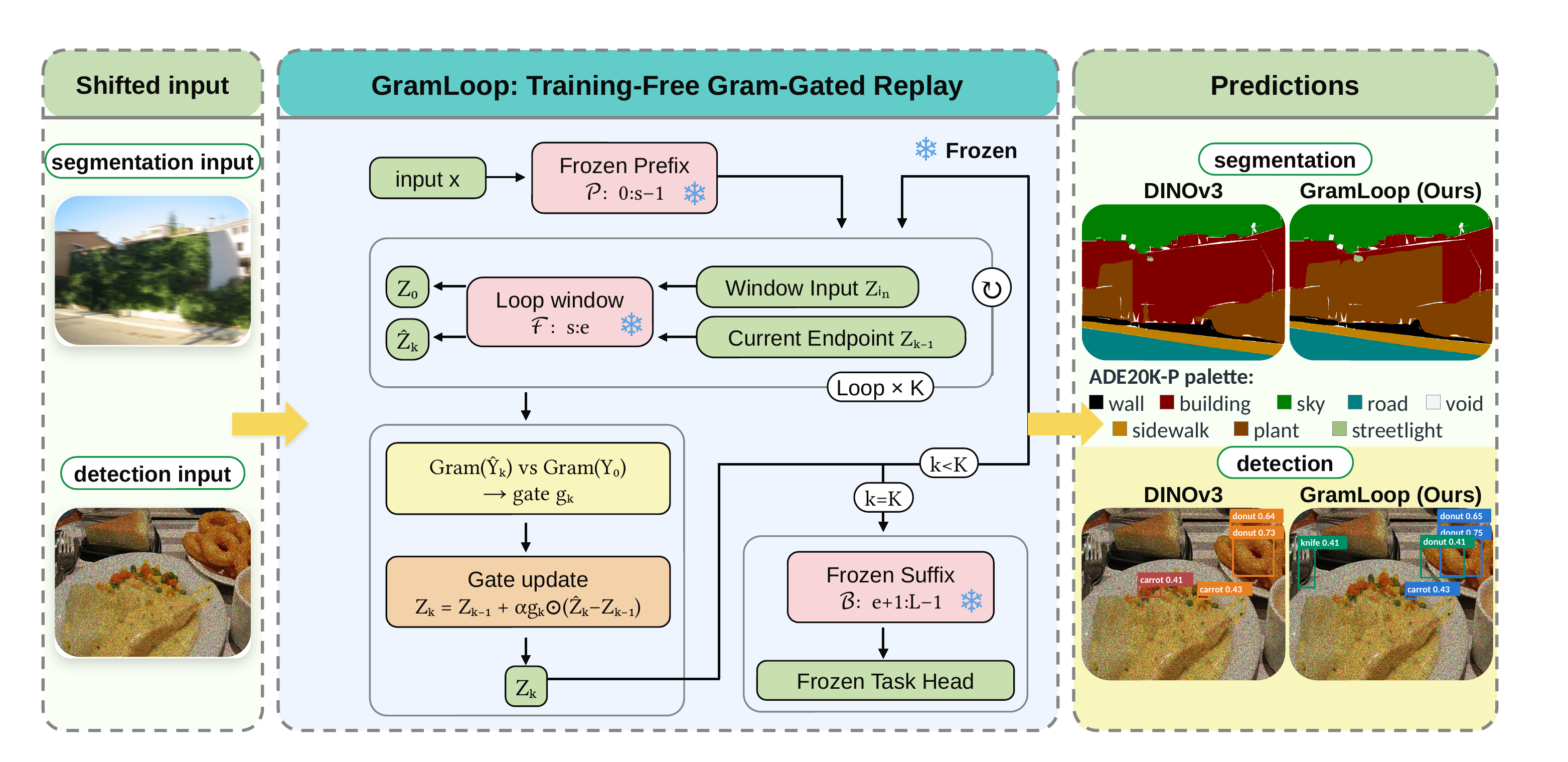}
  \caption{
    \method{} measures the final-layer Gram drift of each replay proposal
    and attenuates structurally disruptive endpoint updates before the unchanged
    task head.
  }
  \label{fig:teaser}
\end{figure*}

\section{Method}
\label{sec:method}

We introduce \method, a training-free Gram-gated replay method for dense
prediction with a frozen DINOv3 backbone. We first review the Gram structure of
DINOv3 patch features, then use that structure to score and gate feature
proposals generated by replaying a short window of transformer blocks. Figure~\ref{fig:teaser}
summarizes how replay proposals, suffix evaluation, Gram-gated endpoint updates, and the
unchanged task head fit together.

\begin{figure}[t]
	\centering
	\includegraphics[width=\columnwidth]{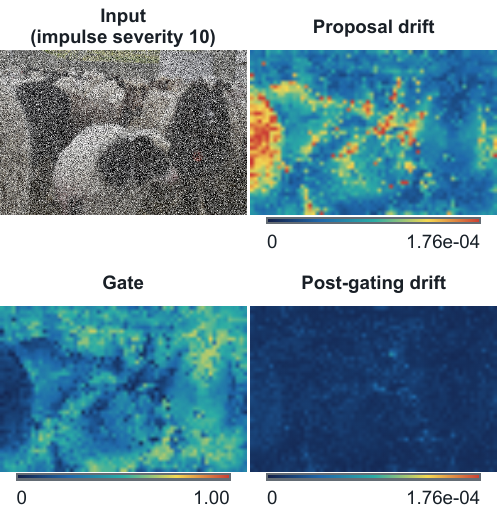}
	\caption{
		Replay proposes patch updates with spatially uneven Gram drift; \method{}
		assigns low gates to high-drift patches, yielding accepted updates with
		lower residual relation error. For a fixed COCO-P exemplar at impulse
		severity 10, the \(2\times2\) panels show the corrupted input, proposal
		drift, gate, and post-gating drift in reading order. Proposal and
		post-gating drift share a color scale, while the gate uses \([0,1]\).
	}
	\label{fig:gram_gated_impulse_diagnostic}
\end{figure}

\subsection{Background: DINO and Gram-Structured Dense Features}
\label{sec:dinov3_gram}

Building on DINO's self-distillation paradigm \citep{caron2021dino}, DINOv3~\citep{simeoni2026dinov3} is a
self-supervised vision backbone designed to produce general-purpose features
for both image-level and dense prediction tasks. A
key challenge in scaling such pretraining is that global representation quality
can continue to improve while the spatial organization of patch features
degrades. DINOv3 addresses this failure mode with \emph{Gram anchoring}: during
training, a student feature map is regularized toward the pairwise
patch-similarity structure produced by an earlier model snapshot
\citep{simeoni2026dinov3}.

Let $\mathbf{q}_p\in\mathbb{R}^{C}$ be the channel-wise normalized feature
vector of patch $p$, where $C$ is the feature-channel dimension, and stack the
$N$ patch vectors as
$\mathbf{Q}=[\mathbf{q}_1,\ldots,\mathbf{q}_N]^{\top}
\in\mathbb{R}^{N\times C}$. Its cosine Gram matrix is
\begin{equation}
    \mathbf{G}(\mathbf{Q}) = \mathbf{Q}\mathbf{Q}^{\top}
    \in\mathbb{R}^{N\times N},
    \label{eq:gram_background}
\end{equation}
where the scalar entry
$G_{pq}:=[\mathbf{G}(\mathbf{Q})]_{pq}=\mathbf{q}_p^{\top}\mathbf{q}_q$
measures the relation between patches $p$ and $q$. Unlike a
pointwise feature penalty, a Gram objective preserves the geometry among all
patches: individual features may change, provided that their relational
structure remains consistent. Matching this matrix therefore directly
constrains the spatial relations that dense prediction relies on without
requiring feature coordinates to remain unchanged. Preserving Gram-structured patch relations
is therefore important for representation stability and downstream dense-prediction performance
\citep{simeoni2026dinov3}.

\subsection{\method}

\paragraph{Motivation}
Distribution shifts can degrade the dense representations produced by a frozen
DINOv3 backbone. We therefore replay a short window of transformer blocks to
refine these features with additional inference-time computation. However,
naively replaying the window updates patch tokens without considering whether
the proposed changes preserve their relational structure and can consequently
disrupt Gram consistency. To control this effect, we treat the standard DINOv3
forward pass as an image-specific relational anchor and measure each replay
proposal by its final-layer Gram deviation from this anchor. Proposals that remain
consistent with the anchor receive larger updates, whereas those with large Gram
deviations are downweighted. This produces Gram-gated replay without a teacher,
labels, gradient updates, or additional training.

Figure~\ref{fig:gram_gated_impulse_diagnostic} illustrates this motivation
at impulse severity 10. High-drift regions receive smaller gates, while the
post-gating drift remains small on the shared drift color scale.

\paragraph{Window Replay and Final-Layer Guidance}
\label{sec:window_loop}
Consider a frozen DINOv3 encoder composed of $L$ transformer blocks
$f_0,\ldots,f_{L-1}$ and a frozen task head $h$. We choose a short loop window
$[s,e]$, where $s$ and $e$ are its start and endpoint block indices. For block
indices $0\leq a\leq b\leq L-1$, define
$\mathcal{B}_{a:b}=f_b\circ\cdots\circ f_a$. We then decompose the encoder into
a prefix, the looped window, and a suffix:
\begin{equation}
    \mathcal{P}=\mathcal{B}_{0:s-1},\qquad
    \mathcal{F}=\mathcal{B}_{s:e},\qquad
    \mathcal{S}=\mathcal{B}_{e+1:L-1}.
    \label{eq:encoder_decomposition}
\end{equation}
Given an image $\mathbf{x}$, the standard forward pass first produces the
window input $\mathbf{Z}_{\mathrm{in}}=\mathcal{P}(\mathbf{x})$ and the endpoint anchor
\begin{equation}
    \mathbf{Z}_0 = \mathcal{F}(\mathbf{Z}_{\mathrm{in}}).
    \label{eq:endpoint_anchor}
\end{equation}
This initial pass through the window is part of the ordinary backbone trajectory
and is not counted as a replay. Propagating the anchor through the suffix gives
the fixed future reference $\mathbf{Y}_0=\mathcal{S}(\mathbf{Z}_0)$. Each endpoint state
$\mathbf{Z}_k\in\mathbb{R}^{T\times C}$ contains all $T$ encoder tokens with
$C$ channels.

Starting from $\mathbf{Z}_{k-1}$, replay step
$k\in\{1,\ldots,K\}$, where $K$ is the prescribed number of additional
replays, forms a proposal by applying the same frozen window again,
\begin{equation}
    \widehat{\mathbf{Z}}_k = \mathcal{F}(\mathbf{Z}_{k-1}),
    \qquad
    \widehat{\mathbf{Y}}_k = \mathcal{S}(\widehat{\mathbf{Z}}_k).
    \label{eq:replay_proposal}
\end{equation}
The temporary suffix pass exposes the proposal's downstream consequence. We
therefore \emph{measure} structural drift at the final probe
$\widehat{\mathbf{Y}}_k$, but \emph{apply} the resulting correction at the
earlier endpoint $\widehat{\mathbf{Z}}_k$. This separation lets final-layer
evidence guide the earlier update: a local proposal is accepted only to the
extent that its mature representation remains compatible with the standard
DINOv3 trajectory.

\paragraph{Patchwise Gram-Consistency Gate}
\label{sec:gram_gate}

\begin{table*}[t]
	\centering
	\begingroup
	\small
	\setlength{\tabcolsep}{3pt}
	\renewcommand{\arraystretch}{1.08}
	\begin{tabular}{@{}lcccc@{\hspace{8pt}}|ccc@{}}
		\toprule
		& \multicolumn{4}{c}{\shortstack{\textbf{Object Detection}\\Plain-DETR~\citep{lin2023plaindetr}}}
		& \multicolumn{3}{c}{\shortstack{\textbf{Semantic Segmentation}\\Mask2Former~\citep{cheng2022mask2former}}} \\
		\cmidrule(lr){2-5}\cmidrule(lr){6-8}
		& \multicolumn{2}{c}{\textbf{COCO-O}}
		& \textbf{COCO-P}
		& \textbf{COCO-C}
		& \textbf{ADE20K}
		& \textbf{ADE20K-P}
		& \textbf{ADE20K-C} \\
		\cmidrule(lr){2-3}\cmidrule(lr){4-4}\cmidrule(lr){5-5}
		\cmidrule(lr){6-6}\cmidrule(lr){7-7}\cmidrule(lr){8-8}
		\textbf{Method}
		& \textbf{mAP \up} & \textbf{ER \up}
		& \textbf{\shortstack{mPC\\(AP) \up}}
		& \textbf{\shortstack{mPC\\(AP) \up}}
		& \textbf{mIoU \up}
		& \textbf{\shortstack{mPC\\(mIoU) \up}}
		& \textbf{\shortstack{mPC\\(mIoU) \up}} \\
		\midrule
		DINOv3~\citep{simeoni2026dinov3}
		& 66.264 & 36.746
		& 59.146
		& 54.901
		& 62.553
		& 59.003
		& 56.558 \\
		\midrule
		\rowcolor{gray!10}
		\method{} (Ours)
		& \best{66.516}
		& \best{36.996}
		& \best{59.205}
		& \best{54.986}
		& \best{62.717}
		& \best{59.040}
		& \best{56.611} \\
		\rowcolor{gray!10}
		$\Delta$
		& \textcolor{\deltaPosColor}{+0.252}
		& \textcolor{\deltaPosColor}{+0.250}
		& \textcolor{\deltaPosColor}{+0.060}
		& \textcolor{\deltaPosColor}{+0.085}
		& \textcolor{\deltaPosColor}{+0.165}
		& \textcolor{\deltaPosColor}{+0.037}
		& \textcolor{\deltaPosColor}{+0.053} \\
		\bottomrule
	\end{tabular}
	\endgroup
	\caption{
        Training-free \method{} improves all shifted benchmarks across object
        detection and semantic segmentation while preserving clean ADE20K
        performance.
	}
	\label{tab:main_results}
\end{table*}

In detail, we extract the valid patch tokens from $\mathbf{Y}_0$ and
$\widehat{\mathbf{Y}}_k$ after the encoder's final normalization, then normalize
each token along the channel dimension. Denote the resulting matrices by
$\mathbf{Q}_0,\widehat{\mathbf{Q}}_k\in\mathbb{R}^{N\times C}$, where $N$ is
the number of image-grid patches used by the gate. The backbone input contains no batch-padding tokens, so every
produced patch token is valid and $N=P$. The implementation additionally
accepts an explicit caller-provided valid-patch mask for padded or packed
inputs, but this optional path is not active in the reported experiments. We
compute exact global cosine Gram matrices
\begin{equation}
    \mathbf{G}_0=\mathbf{Q}_0\mathbf{Q}_0^{\top},
    \qquad
    \widehat{\mathbf{G}}_k=\widehat{\mathbf{Q}}_k
    \widehat{\mathbf{Q}}_k^{\top}.
    \label{eq:loop_grams}
\end{equation}
For replay $k$, we define the patch-drift vector
$\mathbf{d}_k\in\mathbb{R}_{\geq 0}^{N}$. Its component for valid patch $p$ is
the mean squared change of the corresponding Gram row,
\begin{equation}
    [\mathbf{d}_k]_p = \frac{1}{N}\sum_{q=1}^{N}
    \left(
    [\widehat{\mathbf{G}}_k]_{pq}
    -[\mathbf{G}_0]_{pq}
    \right)^2.
    \label{eq:patch_drift}
\end{equation}
This score measures how much the proposal changes patch $p$'s relation to the
entire image, rather than only its feature vector. To make the gate insensitive
to the absolute drift scale of each image and replay, we normalize by the median
valid-patch drift and map the result componentwise to the patch-gate vector
$\mathbf{g}_k^{\mathrm{patch}}\in(0,1]^N$:
\begin{equation}
    [\mathbf{g}_k^{\mathrm{patch}}]_p=\exp\!\left(
    -\frac{[\mathbf{d}_k]_p}
    {\max\!\left(
    \operatorname{median}(\mathbf{d}_k),\epsilon
    \right)}
    \right).
    \label{eq:exponential_gate}
\end{equation}
Here $\epsilon=10^{-6}$ prevents division by zero when the median valid-patch
drift is zero or numerically negligible.
Intuitively, consistent proposals receive larger updates, whereas proposals
that strongly distort the anchor's relational geometry are suppressed smoothly.
Each scalar component $[\mathbf{g}_k^{\mathrm{patch}}]_p\in(0,1]$ is
deterministic and proposal-specific: it is recomputed for every valid patch and
every replay, rather than learned or shared across images.

\paragraph{Gated Endpoint Update}
\label{sec:gated_update}

Each patch token $p$ receives the scalar gate
$[\mathbf{g}_k^{\mathrm{patch}}]_p$ from
Eq.~\eqref{eq:exponential_gate}. For the non-spatial tokens, we define one
image-level gate
\begin{equation}
    g_k^{\mathrm{sp}}=\frac{1}{N}\sum_{p=1}^{N}
    [\mathbf{g}_k^{\mathrm{patch}}]_p,
    \label{eq:special_token_gate}
\end{equation}
and assign it to the class token and each of the $R$ register tokens. The
resulting full token gate is
$\mathbf{g}_k=[g_k^{\mathrm{sp}},\ldots,g_k^{\mathrm{sp}},
[\mathbf{g}_k^{\mathrm{patch}}]_1,\ldots,
[\mathbf{g}_k^{\mathrm{patch}}]_P]^\top\in(0,1]^T$, where the first $R+1$ entries
correspond to the class and register tokens. Its entries are broadcast over
the $C$ channels. We interpolate toward the proposal at the replay-window
endpoint using the tokenwise gate:
\begin{equation}
    \mathbf{Z}_k = \mathbf{Z}_{k-1}
    + \mathbf{g}_k\odot
    \left(\widehat{\mathbf{Z}}_k-\mathbf{Z}_{k-1}\right),
    \label{eq:gated_endpoint_update}
\end{equation}
where $\odot$ denotes tokenwise multiplication with channel broadcasting.
Equation~\eqref{eq:gated_endpoint_update} is a per-token interpolation: a gate
near zero retains the current state, while a larger gate accepts more of the
proposal displacement. Together, Eqs.~\eqref{eq:exponential_gate}
and~\eqref{eq:gated_endpoint_update} show that, within a replay, the accepted
displacement of patch $p$ decreases exponentially with its final-layer Gram
drift $[\mathbf{d}_k]_p$. High-drift proposals therefore produce smaller
endpoint changes. This mechanism modulates replay updates rather than
optimizing a Gram loss or forcing the replayed representation to equal the
anchor.
For token $t$, we denote the corresponding proposal and current-state vectors by
$\widehat{\mathbf{z}}_{k,t}:=[\widehat{\mathbf{Z}}_k]_{t,:}$ and
$\mathbf{z}_{k-1,t}:=[\mathbf{Z}_{k-1}]_{t,:}$, respectively. The class and
register tokens share only the scalar $g_k^{\mathrm{sp}}$: each token follows
its own proposal displacement
$\widehat{\mathbf{z}}_{k,t}-\mathbf{z}_{k-1,t}$ and is therefore updated to a
distinct value. The accepted state $\mathbf{Z}_k$ then becomes the input to
replay $k+1$, so both the proposal and its gate are updated recurrently.

After $K$ additional replays, we recompute the frozen suffix from the accepted
endpoint state and predict with the unchanged task head:
\begin{equation}
    \widehat{\mathbf{o}} = h\!\left(\mathcal{S}(\mathbf{Z}_K)\right).
    \label{eq:final_prediction}
\end{equation}
Here $\mathcal{S}=\mathcal{B}_{e+1:L-1}$ is the frozen encoder suffix from
Eq.~\eqref{eq:encoder_decomposition}, $h$ is the frozen task-specific dense
prediction head, and $\widehat{\mathbf{o}}$ denotes its semantic-segmentation
logits or object-detection outputs. Recomputing $\mathcal{S}$ ensures that the final
prediction follows one coherent backbone trajectory rather than using the
temporary probe features directly.
In all dense-prediction instantiations, \method{} changes only this backbone
token trajectory before the unchanged task-specific adapter and head.
 \section{Experiments}

\begin{table}[t]
	\centering
	\begingroup
	\small
	\setlength{\tabcolsep}{4pt}
	\begin{tabular}{@{}lcc@{}}
		\toprule
		\textbf{Benchmark}
		& \textbf{$\Delta$}
		& \textbf{95\% CI} \\
		\midrule
		ADE20K-C & +0.053 & $[+0.027,+0.082]$ \\
		ADE20K-P & +0.037 & $[+0.015,+0.059]$ \\
        \midrule
		COCO-C   & +0.085 & $[+0.057,+0.114]$ \\
		COCO-P   & +0.060 & $[+0.040,+0.079]$ \\
		\bottomrule
	\end{tabular}
	\endgroup
	\caption{
		All four shifted-benchmark gains of \method{} have 95\% confidence
		intervals that exclude zero under paired cluster bootstrapping. Intervals
		use 10,000 replicates (seed 20260724) and exclude clean conditions.
	}
	\label{tab:cluster_bootstrap_ci}
\end{table}

\subsection{Datasets and Evaluation Benchmarks}
\label{sec:datasets_evaluation}

\paragraph{Distribution-Shift Benchmarks.}
We evaluate \method{} under three complementary forms of distribution shift:
controlled corruptions, real-world-inspired perturbations, and natural domain
shifts. The -C suites cover 15 common corruptions
\citep{hendrycks2019corruptions,kamann2021segmentationcorruptions,michaelis2019robustness};
the -P suites cover 17 real-world-inspired perturbations
\citep{schiappa2024robustness}; and COCO-O spans six natural visual domains
\citep{mao2023cocoo}. We instantiate these shift regimes on two
dense-prediction tasks: ADE20K \citep{zhou2017ade20k} for semantic segmentation
and COCO \citep{lin2014coco} for object detection. This design tests whether
loop-based feature refinement remains effective across distinct sources of
shift while the released DINOv3 backbone, task adapters, and prediction heads
remain fixed. Each shifted condition is paired with its clean task setting.

\paragraph{Semantic segmentation.}
ADE20K SceneParse150 provides 2,000 validation images and 150 object and stuff
classes \citep{zhou2017ade20k}. ADE20K-C applies 15 label-preserving corruptions
at five severity levels
\citep{hendrycks2019corruptions,kamann2021segmentationcorruptions}, and the
released ADE20K-P benchmark applies 17 perturbations at five severity levels
\citep{schiappa2024robustness}.

\paragraph{Object detection.}
COCO val2017 contains 5,000 images and 80 bounding-box categories
\citep{lin2014coco}. COCO-C applies 15 corruptions at five severity levels
\citep{michaelis2019robustness}, COCO-P provides 17 perturbations at five
severity levels with paired box annotations \citep{schiappa2024robustness}, and
COCO-O contains 6,782 images from six natural-shift domains
\citep{mao2023cocoo}.

\begin{figure}[t]
	\centering
	\includegraphics[width=\columnwidth]{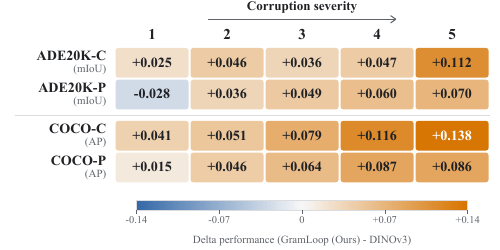}
	\caption{
		The gains from \method{} generally grow with shift severity, and each
		benchmark's largest gain occurs at severity 4 or 5. Columns correspond
		to severity levels 1--5 from left to right.
	}
	\label{fig:severity_heatmap}
\end{figure}

\subsection{Experimental Metrics}
\label{sec:evaluation_metrics}

For semantic segmentation, we report mean intersection over union (mIoU).
For the -C and -P benchmarks, mean
performance under corruption (mPC) is the unweighted mean of the corresponding
task metric over all non-clean shift--severity conditions. For COCO-O, mAP is
the unweighted mean of the domain-level AP scores across its six domains. We
additionally report Effective Robustness,
$\mathrm{ER}=\mathrm{AP}_{\mathrm{COCO\text{-}O}
}-0.45\,\mathrm{AP}_{\mathrm{COCO}}$, following the clean-performance
correction of \citet{mao2023cocoo}. All metrics are reported in percentage
points, and higher values are better. DINOv3 and \method{} are evaluated on
identical images and annotations.

\subsection{Experimental Details}
\label{sec:implementation_details}

Unless otherwise stated, we use the 40-block DINOv3 ViT-7B/16 backbone
\citep{simeoni2026dinov3}. \method{} replays blocks $[21,23]$ for $K=2$
additional replays and measures each proposal's cosine-Gram drift at the final
block.

We retain the released task interfaces: ViT-Adapter with Mask2Former for
ADE20K \citep{chen2023vitadapter,cheng2022mask2former} and Plain-DETR for COCO
\citep{lin2023plaindetr}. Both heads consume spatial patch features from the
same four taps. Evaluation is
deterministic with seed 100. All backbone, adapter, and task-head parameters
remain fixed, so \method{} is training-free.
The supplementary material provides additional details on token routing,
preprocessing, valid-token handling, and shifted-benchmark annotations. Our code will be released.

\subsection{Experimental Results}

\subsubsection{Main Results}

\begin{figure*}[t]
	\centering
	\includegraphics[width=0.72\textwidth]{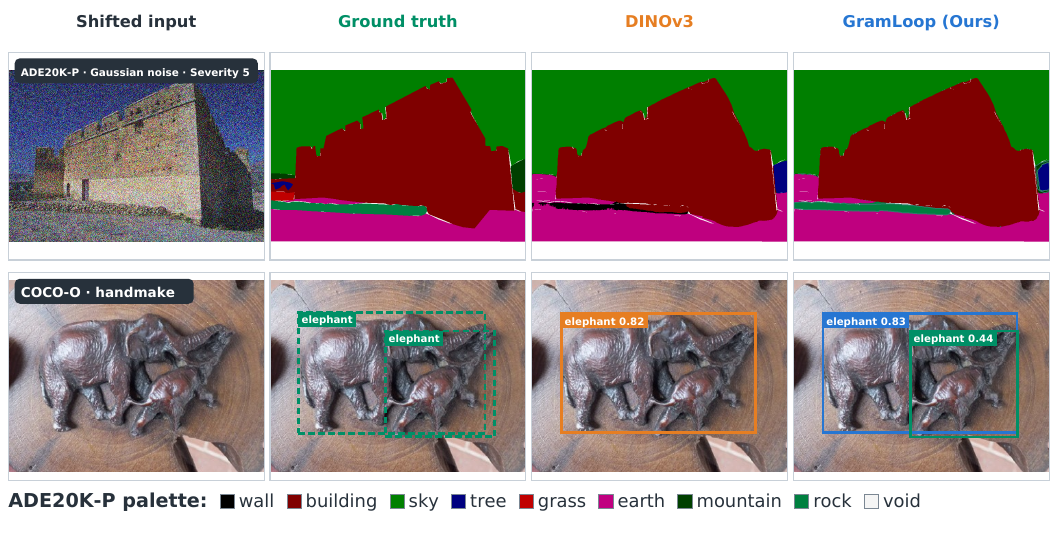}
	\caption{
		\method{} corrects a large semantic region and recovers a missed object
		across segmentation and natural detection shifts. Detection labels report
		confidence scores using a shared threshold of $0.40$. The ADE20K-P
		example uses severity 5, whereas COCO-O handmake is a natural-shift
		domain.
	}
	\label{fig:qualitative_shift_examples}
\end{figure*}

\begin{table}[tb]
	\centering
	\small
	\setlength{\tabcolsep}{1.2pt}
	\renewcommand{\arraystretch}{1.05}
	\begin{tabular}{@{}lcccc@{}}
		\toprule
		\textbf{Method }
		& \textbf{$\mathcal{S}_1$}
		& \textbf{$\mathcal{S}_2$}
		& \textbf{$\mathcal{S}_3$}
		& \textbf{$\mathcal{S}_4$} \\
		\midrule
		\multicolumn{5}{@{}l}{\color{gray}\textit{Fixed subset} $\mathcal{S}_i$} \\
		DINOv3
		& 60.070 & 60.828 & 60.026 & 59.913 \\
		\method{}
		& \best{60.424}
		& \best{60.684}
		& \best{60.105}
		& \best{59.965} \\
		$\Delta$
		& \textcolor{\deltaPosColor}{+0.354}
		& \textcolor{\deltaNegColor}{-0.144}
		& \textcolor{\deltaPosColor}{+0.079}
		& \textcolor{\deltaPosColor}{+0.052} \\
		\addlinespace[1pt]
        \midrule
        \midrule
		\multicolumn{5}{@{}l}{\color{gray}\textit{Complement} $\mathcal{V}\setminus\mathcal{S}_i$} \\
		DINOv3
		& 62.299 & 62.236 & 62.101 & 62.474 \\
        \rowcolor{gray!10}
		\method{}
		& \best{62.325}
		& \best{62.582}
		& \best{62.250}
		& \best{62.603} \\
        \rowcolor{gray!10}
		$\Delta$
		& \textcolor{\deltaPosColor}{+0.026}
		& \textcolor{\deltaPosColor}{+0.346}
		& \textcolor{\deltaPosColor}{+0.149}
		& \textcolor{\deltaPosColor}{+0.129} \\
		\bottomrule
	\end{tabular}
	\caption{
		\method{} with fixed cross-validation hyperparameters improves three of
		four ADE20K subsets and all four complements.
	}
	\label{tab:ade20k_partition_consistency}
\end{table}

\paragraph{Training-free \method{} improves shifted dense-prediction robustness.}
Across COCO-O, COCO-P, COCO-C, ADE20K-P, and ADE20K-C,
Table~\ref{tab:main_results} reports the main comparison. \method{} improves
over the paired DINOv3 baseline on every shifted benchmark. Although the gains
on the -C and -P suites are modest, they are obtained without training or
parameter updates on a frozen 7B-parameter DINOv3 backbone pretrained on
1.7 billion images \citep{simeoni2026dinov3}. These consistent gains are
therefore nontrivial. Improvements across both Plain-DETR detection and
Mask2Former segmentation further indicate that the gains arise from refining
the shared token trajectory rather than modifying task-specific heads.
\method{} also improves clean ADE20K by \(+0.165\) mIoU.
Figure~\ref{fig:qualitative_shift_examples} complements these aggregate scores:
on ADE20K-P, \method{} corrects a large semantic region rather than only
scattered pixels; on COCO-O handmake, it recovers a missed object at the same
\(0.40\) score threshold. Because the task heads and inference hyperparameters
remain fixed, these corrections reflect refined backbone evidence rather than
retuning the downstream predictor. The gains are also broadly distributed
across semantic categories: 94 of 150 ADE20K classes improve when class IoU is
averaged over all 75 ADE20K-C conditions. This majority-class coverage shows
that the aggregate improvement is not concentrated in a few labels; the
supplementary material lists the 10 largest gains.

\paragraph{The improvements persist under resampling.}
To assess whether the aggregate gains are driven by only a few shift families,
we perform paired cluster bootstrapping over corruption or perturbation
families. Each replicate retains all five severity levels of every sampled
family and recomputes the paired mean difference. As shown in
Table~\ref{tab:cluster_bootstrap_ci}, all four 95\% confidence intervals exclude
zero, supporting the consistency of the improvements across shift families.

\paragraph{The gains generally increase with shift severity.}
After establishing the aggregate improvements and their bootstrap consistency,
we examine how the gains vary with shift severity.
Figure~\ref{fig:severity_heatmap} reports the paired performance change at each
severity level, averaged over the corresponding shift families. Across the four
suites, the gains generally concentrate at severities 4--5, indicating that
selective replay is most beneficial under stronger distribution shifts.

\paragraph{Fixed cross-validation hyperparameters transfer across validation partitions.}
Let $\mathcal{V}$ denote the complete 2,000-image ADE20K validation set, and let
$\mathcal{S}_i$ denote its $i$th fixed 500-image subset.
Table~\ref{tab:ade20k_partition_consistency} reports mIoU on each
$\mathcal{S}_i$ and on its 1,500-image complement
$\mathcal{V}\setminus\mathcal{S}_i$. The fixed subsets improve in three of
four cases, and all complements improve, indicating that the aggregate mIoU
gain is not driven by a single validation partition.

\subsubsection{Ablations and Diagnostics}

\paragraph{Final-layer Gram gating with loop-end updates is the best ablation.}
Table~\ref{tab:placement_ablation} shows that measuring Gram consistency at the
final layer while applying the accepted update at the loop endpoint performs
best. This result supports separating reliability estimation from trajectory
modification.

\begin{table}[t]
	\centering
	\begingroup
	\small
	\setlength{\tabcolsep}{3pt}
	\begin{tabular}{@{}lc|cc@{}}
		\toprule
		\textbf{Gate at}
		& \textbf{Update at}
		& \textbf{mAP \up}
		& \textbf{ER \up} \\
		\midrule
		-- & -- & 66.264 & 36.746 \\
		\midrule
		Loop end & Loop end & 66.369 & 36.862 \\
		Final layer & Final layer & 66.174 & 36.705 \\
		\rowcolor{gray!10}
		Final layer & Loop end & \best{66.516} & \best{36.996} \\
		\bottomrule
	\end{tabular}
	\endgroup
	\caption{
		Final-layer Gram evidence performs best when the accepted update is
		applied at the loop endpoint. ``Gate at'' and ``Update at'' denote where the gate is
    computed and applied; ``Loop end'' and ``Final layer'' are zero-indexed
    blocks 23 and 39, respectively.
	}
	\label{tab:placement_ablation}
\end{table}

\begin{figure}[t]
	\centering
	\includegraphics[width=\columnwidth]{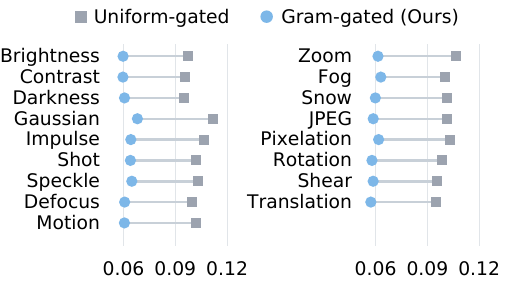}
	\caption{
		At matched mean update strength, patchwise Gram gating lowers final-layer
		cosine-Gram discrepancy for all 17 COCO-P perturbations. Discrepancies are
		normalized by Ungated (\(=1\)); lower is better.
	}
	\label{fig:gram_consistency}
\end{figure}

\noindent\begin{minipage}{\columnwidth}
\paragraph{Patchwise gating preserves Gram consistency.}
At matched mean update strength, every patchwise-gated marker lies left of its
uniform-gated counterpart across all 17 COCO-P perturbations and 85 severity
conditions (Figure~\ref{fig:gram_consistency}). Thus, token selection matters
beyond taking a smaller replay step.
Table~\ref{tab:compute_cost} reports \(1.96\times\) GPU-equivalent inference
time with unchanged peak memory, making runtime the main trade-off.
\end{minipage}

\begin{table}[!b]
    \centering
    \begingroup
    \small
    \setlength{\tabcolsep}{2.2pt}
    \begin{tabular}{@{}l|ccccc@{}}
        \toprule
        \textbf{Path}
        & \textbf{Blocks}
        & \textbf{Analytic}
        & \textbf{\shortstack{GPU time\\(s/image) \down}}
        & \textbf{Obs.}
        & \textbf{\shortstack{Peak\\GB \down}} \\
        \midrule
        Baseline & 40 & \(1.00\times\) & 2.73 & \(1.00\times\) & 46.56 \\
        Endpoint replay
        & 46 & \(1.15\times\) & 3.14
        & \(1.15\times\) & 46.56 \\
        Final filter
        & 78 & \(1.95\times\) & 4.63 & \(1.70\times\) & 46.56 \\
        \hline
        \rowcolor{gray!10}
        \method{} (Ours)
        & 94 & \(2.35\times\) & 5.35 & \(1.96\times\) & 46.56 \\
        \bottomrule
    \end{tabular}
    \endgroup
    \caption{
        Measured inference cost remains below the block-count estimate, while
        peak GPU memory is unchanged. ``Endpoint replay,'' ``Final-state filtering,'' and \method{} use gate/update
locations of loop-end/loop-end, final-layer/final-layer, and
final-layer/loop-end, respectively.
    }
    \label{tab:compute_cost}
\end{table}

\FloatBarrier
 \section{Conclusion}

We introduced \method{}, a training-free method for improving the robustness of
frozen DINOv3 dense-prediction models. At inference time, it replays a short
transformer window and uses final-layer Gram consistency to gate patchwise
updates, while leaving all model parameters and task heads unchanged. Across
COCO-O, COCO-P, COCO-C, ADE20K-P, and ADE20K-C, \method{} improves over the
paired DINOv3 baseline while preserving clean ADE20K performance.
Ablations identify final-layer global Gram evidence applied at the replay-window
endpoint as the strongest evaluated configuration. For the future work, we can reduce inference
time by adaptively determining when and how many times to replay and by developing
lower-cost reliability probes, while preserving the gains under distribution shifts.

\section{Acknowledgement}
We thank Yuki Asano for the discussion in this work.

\bibliography{references}

\clearpage
\appendix
\newcommand{\suppmainfigureref}[2]{Figure~\ref{#1}}

\captionsetup{font=small}

\section{Supplementary Material}

\subsection{Supplementary Method Details}
\label{app:method_details}

\begin{algorithm}[H]
	\caption{\textbf{Training-free Gram-gated replay.}
		$\operatorname{Patch}$ selects valid patch tokens,
		$\operatorname{RowNorm}_2$ normalizes each patch-token feature to unit
		$\ell_2$ norm, and $\operatorname{RowMSE}$ returns the mean squared
		discrepancy of each Gram row. The prefix and suffix retain the tapped
		features required by the dense-prediction head.}
	\label{alg:gramloop}
	\begin{algorithmic}[1]
		\small
		\Require Image $\mathbf{x}$; frozen tapped prefix $\mathcal{P}_{\mathrm{tap}}$,
		replay window $\mathcal{F}$, tapped suffix $\mathcal{S}_{\mathrm{tap}}$,
		and task head $h$; replay count $K$; numerical constant $\epsilon>0$
		\Ensure Dense prediction $\widehat{\mathbf{o}}$
		\State $(\mathbf{Z}_{\mathrm{in}},\mathcal{T}_{\mathrm{pre}})
		\gets\mathcal{P}_{\mathrm{tap}}(\mathbf{x})$,
		$\mathbf{Z}_0\gets\mathcal{F}(\mathbf{Z}_{\mathrm{in}})$,
		$(\mathbf{Y}_0,\_)\gets\mathcal{S}_{\mathrm{tap}}(\mathbf{Z}_0)$
		\State $\mathbf{Q}_0\gets
		\operatorname{RowNorm}_2(\operatorname{Patch}(\operatorname{LN}(\mathbf{Y}_0)))$,
		$\mathbf{G}_0\gets\mathbf{Q}_0\mathbf{Q}_0^\top$
		\For{$k=1,\ldots,K$}
		\State $\widehat{\mathbf{Z}}_k\gets\mathcal{F}(\mathbf{Z}_{k-1})$,
		$(\widehat{\mathbf{Y}}_k,\_)\gets
		\mathcal{S}_{\mathrm{tap}}(\widehat{\mathbf{Z}}_k)$
		\State $\widehat{\mathbf{Q}}_k\gets
		\operatorname{RowNorm}_2(
		\operatorname{Patch}(\operatorname{LN}(\widehat{\mathbf{Y}}_k)))$,
		$\widehat{\mathbf{G}}_k\gets
		\widehat{\mathbf{Q}}_k\widehat{\mathbf{Q}}_k^\top$
		\State $\mathbf{d}_k\gets
		\operatorname{RowMSE}(\widehat{\mathbf{G}}_k,\mathbf{G}_0)$
		\State $\mathbf{g}^{\mathrm{patch}}_k\gets
		\exp\!\left(-\mathbf{d}_k/
		\max(\operatorname{median}(\mathbf{d}_k),\epsilon)\right)$
		\State Assign the mean patch gate to the class and register tokens,
		forming the full token gate $\mathbf{g}_k$
		\State $\mathbf{Z}_k\gets\mathbf{Z}_{k-1}
		+\mathbf{g}_k\odot
		(\widehat{\mathbf{Z}}_k-\mathbf{Z}_{k-1})$
		\EndFor
		\State $(\_,\mathcal{T}_{\mathrm{post}})
		\gets\mathcal{S}_{\mathrm{tap}}(\mathbf{Z}_K)$
		\State \Return $\widehat{\mathbf{o}}\gets
		h(\mathcal{T}_{\mathrm{pre}}\cup\mathcal{T}_{\mathrm{post}})$
	\end{algorithmic}
\end{algorithm}

Algorithm~\ref{alg:gramloop} summarizes the complete inference procedure. The
initial forward pass provides a fixed image-specific Gram anchor. At each
replay, the proposal is propagated through the frozen suffix, compared with
this anchor at the final layer, and converted into a patchwise gate that
updates the replay-window endpoint. After $K$ gated replays, the suffix and the
unchanged task head are evaluated once more using the retained multi-level
features to produce the final prediction.

\subsection{Additional Experimental Protocol}
\label{app:additional_protocol}

\paragraph{Evaluation configuration.}
All experiments use the frozen 40-block DINOv3 ViT-7B/16 backbone and released
dense-prediction heads, with zero-indexed taps $[9,19,29,39]$. We replay blocks
$[21,23]$ for $K=2$ additional passes, compute exact global cosine-Gram matrices
at the final layer, apply the gate at block 23, and set $\epsilon=10^{-6}$.
ADE20K uses ViT-Adapter and Mask2Former with $896\times896$ sliding-window
crops; COCO uses Plain-DETR at short-side resolution 2048 with a $3\times3$
local-view grid and one resized global view. No test-time augmentation is used.

\paragraph{Hardware and runtime.}
All evaluations run with batch size one on NVIDIA H100 SXM5 GPUs with 64\,GB
of memory. Parameters, backbone outputs, Gram statistics, and task heads use
\texttt{float32}; backbone block computation, including replay and suffix
passes, uses \texttt{bfloat16} autocasting. No model parameter is updated.

\paragraph{Seed independence.}
The inference path contains no stochastic sampling: models are in evaluation
mode, data order is fixed, and deterministic CUDA algorithms are enforced.
Thus, for fixed inputs, checkpoints, and configuration, predictions do not
depend on the experiment seed; seed 100 is retained only for defensive
random-number-generator initialization and provenance. COCO-C corruptions use
deterministic per-image hashes independent of this experiment seed.

\paragraph{Backbone token routing.}
The class token and four register tokens use the mean unit-range gate over the
patch grid and are updated with their respective proposal displacements.
Because taps 9 and 19 precede the replay window, they retain their
standard-forward features; taps 29 and 39 are recomputed from the accepted
loop endpoint and therefore carry the refined trajectory.

\paragraph{Task interfaces.}
For semantic segmentation, each tapped representation supplies normalized
patch tokens and its class token to the released DINOv3 feature adapter
\citep{simeoni2026dinov3}. Register tokens are not passed to the segmentation
head, and the class token carried by the intermediate-feature interface does
not enter the feature-pyramid computation. Patch tokens are reshaped to their
image grid and fused with the ViT-Adapter convolutional spatial prior
\citep{chen2023vitadapter}; Mask2Former then decodes the resulting four-level
pyramid \citep{cheng2022mask2former}.

For object detection, the released interface retains only normalized patch
tokens from the four taps, reshapes them into spatial maps, and concatenates
the maps along the channel dimension \citep{simeoni2026dinov3}. Class and
register tokens are not passed directly to the detector. Plain-DETR processes
the concatenated representation with its original prediction and
post-processing pipeline \citep{lin2023plaindetr}. Thus patch tokens are the
direct backbone inputs to both task heads; class and register tokens affect
predictions only indirectly through self-attention in the replayed window and
recomputed suffix.

\paragraph{Input and valid-token handling.}
ADE20K is evaluated one sliding-window crop at a time with a nominal
\(896\times896\) crop. COCO is evaluated one image at a time after both spatial
dimensions are resized to multiples of the patch size (16). Neither evaluator
adds batch padding at the backbone input or supplies an external valid-patch
mask. Consequently, every produced patch token is valid and all reported runs
use \(N=P\).

\paragraph{Shifted-benchmark annotations.}
ADE20K-C transforms only the input image, so the original semantic labels are
retained for every corruption--severity condition. COCO-C likewise retains
the original val2017 bounding boxes. MS COCO-P was introduced for segmentation
robustness but releases condition-specific COCO-format
\texttt{instances\_val2017} annotations; we use these provided annotations to
define the paired detection conditions \citep{schiappa2024robustness}.

\paragraph{Extended result breakdowns.}
Tables~\ref{tab:ade20kc_per_corruption} and~\ref{tab:cocop_per_perturbation}
report five-severity means for every ADE20K-C corruption and COCO-P
perturbation, respectively. Table~\ref{tab:ade20kc_top10_classes} further lists
the 10 ADE20K semantic classes with the largest mean gains under ADE20K-C.

\begin{table*}[!t]
    \begin{minipage}[t]{0.47\textwidth}
    \centering
    \begingroup
    \small
    \begin{tabular}{@{}lrrr@{}}
        \toprule
        \textbf{Corruption} & \textbf{DINOv3}
        & \shortstack{\textbf{\method{}}\\\textbf{(Ours)}}
        & \textbf{$\Delta$} \\
        \midrule
        Brightness        & 61.595 & 61.612 & +0.016 \\
        Contrast          & 61.963 & 61.957 & -0.006 \\
        Defocus blur      & 54.126 & 54.100 & -0.026 \\
        Elastic transform & 58.489 & 58.521 & +0.032 \\
        Fog               & 61.364 & 61.374 & +0.010 \\
        Frost             & 53.663 & 53.811 & +0.148 \\
        Gaussian noise    & 55.234 & 55.257 & +0.022 \\
        Glass blur        & 52.759 & 52.892 & +0.133 \\
        Impulse noise     & 57.546 & 57.584 & +0.038 \\
        JPEG compression  & 59.435 & 59.578 & +0.144 \\
        Motion blur       & 55.691 & 55.767 & +0.075 \\
        Pixelate          & 60.097 & 60.125 & +0.028 \\
        Shot noise        & 55.526 & 55.620 & +0.094 \\
        Snow              & 58.488 & 58.480 & -0.008 \\
        Zoom blur         & 42.388 & 42.486 & +0.098 \\
        \midrule
        \textbf{Macro mean} & \textbf{56.558} & \textbf{56.611}
        & \textbf{+0.053} \\
        \bottomrule
    \end{tabular}
    \endgroup
    \caption{Five-severity ADE20K-C mIoU means by corruption; deltas are
    computed before rounding.}
    \label{tab:ade20kc_per_corruption}
    \end{minipage}
    \hfill
    \begin{minipage}[t]{0.47\textwidth}

    \centering
    \begingroup
    \small
    \begin{tabular}{@{}lrrr@{}}
        \toprule
        \textbf{Perturbation} & \textbf{DINOv3}
        & \shortstack{\textbf{\method{}}\\\textbf{(Ours)}}
        & \textbf{$\Delta$} \\
        \midrule
        Gaussian   & 56.228 & 56.342 & +0.114 \\
        Shot       & 62.985 & 63.026 & +0.040 \\
        Impulse    & 58.091 & 58.196 & +0.105 \\
        Speckle    & 62.289 & 62.339 & +0.050 \\
        Defocus    & 45.748 & 45.838 & +0.089 \\
        Motion     & 47.760 & 47.849 & +0.089 \\
        Zoom       & 44.732 & 44.863 & +0.132 \\
        Contrast   & 64.178 & 64.197 & +0.018 \\
        Brightness & 64.186 & 64.201 & +0.015 \\
        Darkness   & 65.483 & 65.500 & +0.018 \\
        JPEG       & 58.916 & 58.988 & +0.072 \\
        Pixelate   & 57.308 & 57.420 & +0.112 \\
        Fog        & 64.463 & 64.466 & +0.003 \\
        Snow       & 60.153 & 60.241 & +0.088 \\
        Rotate     & 63.972 & 64.023 & +0.050 \\
        Translate  & 64.160 & 64.170 & +0.009 \\
        Shear      & 64.822 & 64.830 & +0.008 \\
        \midrule
        \textbf{Macro mean} & \textbf{59.146} & \textbf{59.205}
        & \textbf{+0.060} \\
        \bottomrule
    \end{tabular}
    \endgroup
    \caption{Five-severity COCO-P mAP means by perturbation; deltas are
    computed before rounding.}
    \label{tab:cocop_per_perturbation}
    \end{minipage}
\end{table*}

\begin{table}[ht]
    \centering
    \begingroup
    \small
    \begin{tabular}{@{}c l ccc@{}}
        \toprule
        \textbf{Rank} & \textbf{Class}
        & \textbf{DINOv3}
        & \shortstack{\textbf{\method{}}\\\textbf{(Ours)}}
        & \textbf{$\Delta$} \\
        \midrule
         1 & monitor       & 23.479 & 26.358 & +2.878 \\
         2 & screen door   & 66.339 & 67.028 & +0.689 \\
         3 & airplane      & 81.514 & 82.173 & +0.659 \\
         4 & sculpture     & 67.221 & 67.784 & +0.563 \\
         5 & pot           & 61.488 & 62.036 & +0.548 \\
         6 & ottoman       & 53.116 & 53.610 & +0.493 \\
         7 & barrel        & 65.320 & 65.797 & +0.477 \\
         8 & swivel chair  & 46.620 & 47.045 & +0.426 \\
         9 & bag           & 18.455 & 18.845 & +0.390 \\
        10 & waterfall     & 56.774 & 57.160 & +0.386 \\
        \bottomrule
    \end{tabular}
    \endgroup
    \caption{
        ADE20K-C improvements extend across semantic categories; shown are the
        10 largest mean classwise IoU gains.
    }
    \label{tab:ade20kc_top10_classes}
\end{table}

\subsection{Expanded Qualitative and Diagnostic Results}

\begin{figure*}[!t]
	\centering
	\includegraphics[width=\textwidth]{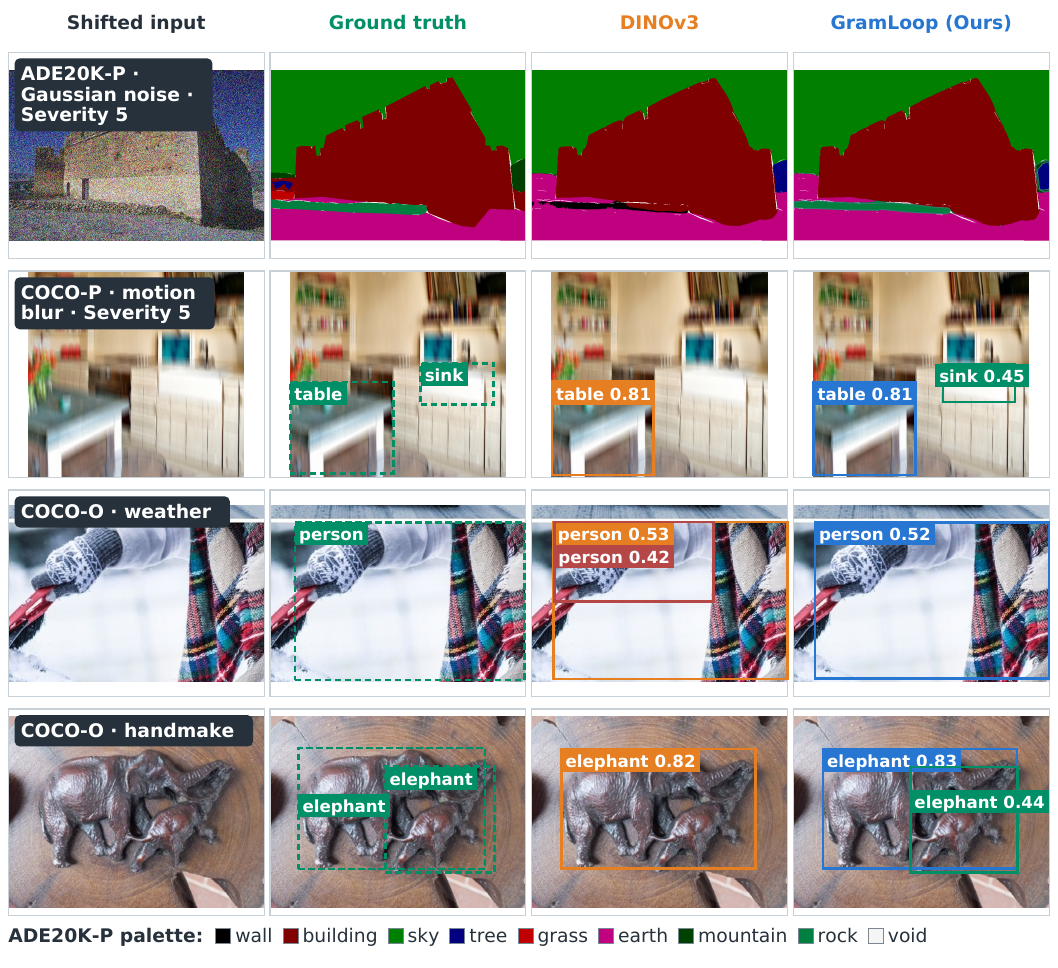}
	\caption{
		Qualitative shifted-input predictions from the same frozen task heads
		used for evaluation. Columns compare the input, annotation, DINOv3, and
		\method{}; rows cover ADE20K-P, COCO-P, and two COCO-O domains.
	}
	\label{fig:qualitative_shift_examples_supp}
\end{figure*}

\begin{figure*}[!t]
	\centering
	\includegraphics[width=\textwidth]{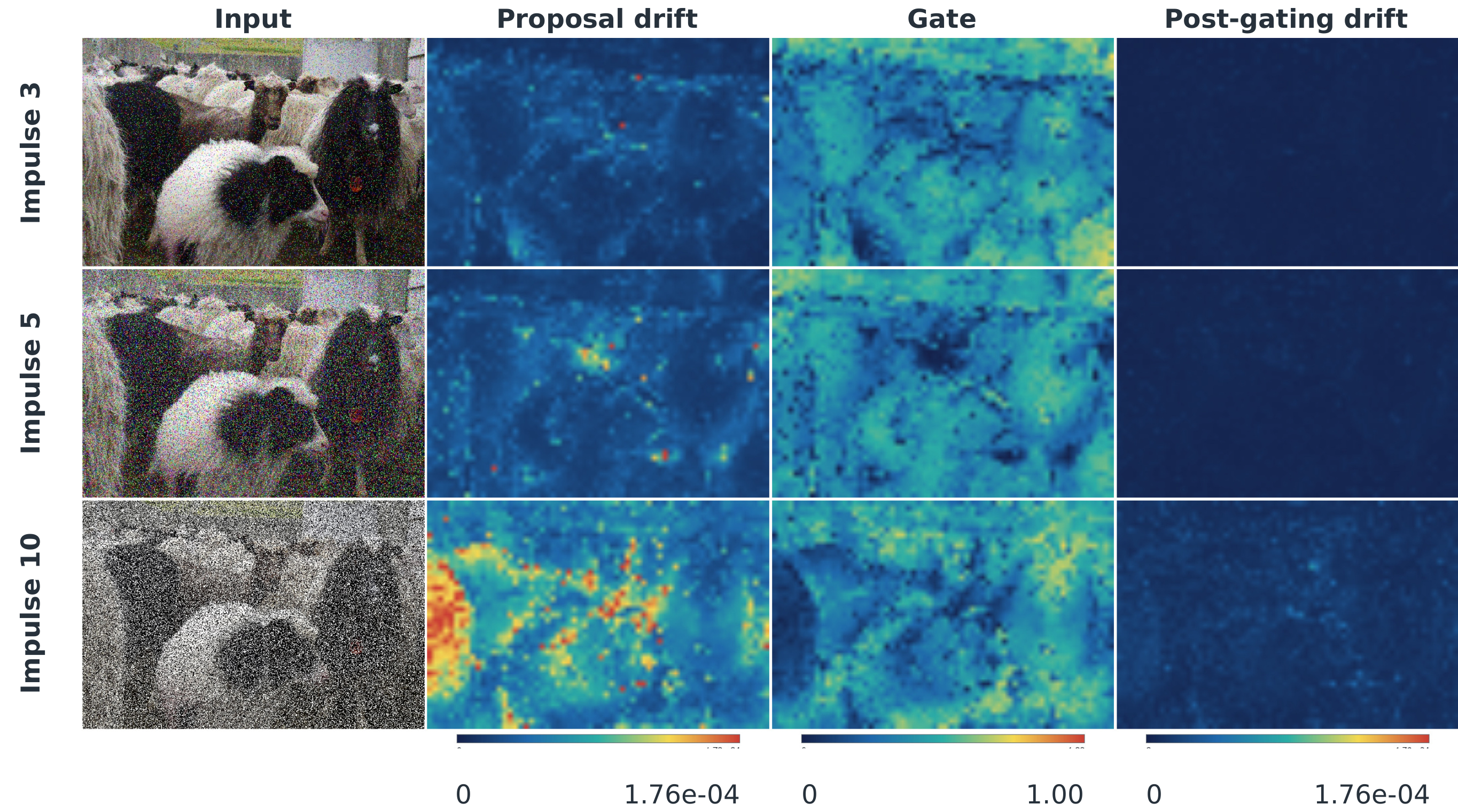}
	\caption{
		Replay-2 diagnostic for COCO image 360325 under increasing impulse
		noise. Columns show the input, proposal drift, gate, and post-gating
		drift.
	}
	\label{fig:gram_gated_impulse_diagnostic_supp}
\end{figure*}

The supplementary figures below extend
\suppmainfigureref{fig:qualitative_shift_examples}{the qualitative comparison
in the main text} and
\suppmainfigureref{fig:gram_gated_impulse_diagnostic}{the impulse-noise
diagnostic in the main text}. Figure~\ref{fig:qualitative_shift_examples_supp}
provides full-width comparisons across ADE20K-P, COCO-P, and two COCO-O
domains, while Figure~\ref{fig:gram_gated_impulse_diagnostic_supp} expands the
diagnostic across multiple impulse-noise severities. Their construction and
visualization protocols are detailed below.

\paragraph{Qualitative-figure construction.}
Every panel is rendered from stored predictions produced by the same frozen
task heads as the quantitative evaluation. For ADE20K-P, the DINOv3 and
\method{} panels are classwise-argmax Mask2Former masks. Ignored annotation
pixels are removed, and the green contour is computed from valid pixels where
\method{} equals the annotation while DINOv3 does not. For detection, each row
uses an identical crop for the input, annotation, and both methods. The display
retains only the illustrated categories and predicted boxes with score at least
\(0.40\) whose centers lie inside that crop. Predictions are sorted by score
and greedily matched to unmatched annotations of the same category at IoU at
least \(0.50\). Dashed green boxes denote annotations; green \method{} boxes
are matched objects not recovered by DINOv3, red DINOv3 boxes are unmatched
detections absent from \method{}, and the remaining predictions retain their
method color. These category, crop, and score filters are used only for
visualization and do not enter the reported metrics.

\paragraph{Impulse-noise diagnostic construction.}
Rows use COCO image 360325. Severities 3 and 5 are released COCO-P inputs;
severity 10 is a deterministic salt-and-pepper extrapolation used only as a
stress test. Let \(\mathbf{G}_0\) be the final-layer cosine-Gram matrix from
the anchor pass and \(\mathbf{G}^{\mathrm{prop}}_2\) the matrix obtained by
propagating the second replay proposal through the frozen suffix. For every
valid patch \(p\), the proposal-drift map is
\[
    r^{\mathrm{prop}}_{2,p}
    = \frac{1}{N}\sum_{q=1}^{N}
    \left(G^{\mathrm{prop}}_{2,pq}-G_{0,pq}\right)^2,
\]
and the gate map is
\[
    g_{2,p}
    = \exp\!\left[
    -\frac{r^{\mathrm{prop}}_{2,p}}
    {\max(\operatorname{median}(\mathbf{r}^{\mathrm{prop}}_2),10^{-6})}
    \right].
\]
The figure displays \(g_{2,p}\), not the complete update coefficient. After
the patchwise gated update is accepted at the loop endpoint, the frozen suffix
is recomputed to obtain \(\mathbf{G}^{\mathrm{gated}}_2\). The final column
displays
\[
    r^{\mathrm{gated}}_{2,p}
    = \frac{1}{N}\sum_{q=1}^{N}
    \left(G^{\mathrm{gated}}_{2,pq}-G_{0,pq}\right)^2.
\]
Each patchwise scalar map is restored to the spatial patch grid. Proposal and
post-gating drift use the same fixed color range, \(0\) to
\(1.76\times10^{-4}\), across all rows, while the gate uses \([0,1]\).

\subsection{Detailed Compute Accounting}
\label{app:compute_accounting}

Let \(B(T,C)\) denote the cost of one transformer block for all \(T\) encoder
tokens and channel width \(C\), \(W=e-s+1\) the replay-window length, and
\(S=L-e-1\) the suffix length. The baseline costs \(LB(T,C)+C_h\), where
\(C_h\) is the unchanged task-head cost. The future-guided endpoint update
costs
\[
    \bigl[L+K(W+S)+S\bigr]B(T,C)+C_h
    +\mathcal{O}\!\left((K+1)N^2C\right).
\]
The final term covers one anchor Gram matrix and one proposal Gram matrix per
replay. Since a transformer block costs
\(\mathcal{O}(TC^2+T^2C)\), \method{} has the same polynomial order as ordinary
inference for fixed \(K\), with an explicit test-time compute multiplier. For
\(L=40\), \(W=3\), \(S=16\), and \(K=2\), the baseline and \method{} execute 40
and 94 equivalent backbone blocks, respectively. The analytic multiplier
counts frozen transformer-block evaluations and excludes the shared task head
and Gram algebra.

For measured timing, GPU-equivalent time per image is the distributed wall
time over the same 11,782 COCO and COCO-O images, multiplied by 64 GPUs and
divided by the image count. It is therefore an accounting quantity rather than
directly measured single-GPU latency. The measured multiplier includes the
task head, Gram computation, and data pipeline, while peak memory is the
maximum observed on any GPU.

\subsection{Additional Gram-Consistency Diagnostics}
\label{app:gram_consistency_diagnostics}

We additionally define the aggregation used for the perturbation-level
Gram-consistency analysis in
\suppmainfigureref{fig:gram_consistency}{the corresponding main-text figure}.
For perturbation \(p\), severity
\(s\), image \(i\), and trajectory
\(m\in\{\mathrm{raw},\mathrm{uniform},\mathrm{gated}\}\), let
\(d_{p,s,i,m}\) denote the mean final-layer cosine-Gram discrepancy. The
perturbation-level normalized value shown there is
\[
    R_{p,m} =
    \frac{\sum_{s,i} d_{p,s,i,m}}
         {\sum_{s,i} d_{p,s,i,\mathrm{raw}}},
    \qquad
    m\in\{\mathrm{uniform},\mathrm{gated}\}.
\]
Here the sums range over all five severities and 5,000 images per severity.
Thus each row aggregates 25,000 paired images before normalization by its
Ungated trajectory.

\end{document}